\documentclass[runningheads]{llncs}

\usepackage{eccv}

\usepackage{eccvabbrv}

\usepackage{graphicx}
\usepackage{booktabs}

\usepackage[accsupp]{axessibility}  

\usepackage{hyperref}

\usepackage{orcidlink}

\begin{document}

\title{DiffSign: AI-Assisted Generation of Customizable \\ Sign Language Videos With Enhanced Realism} 

\titlerunning{DiffSign}

\author{Sudha Krishnamurthy\inst{1} \and
Vimal Bhat\inst{2} \and
Abhinav Jain\inst{2}}

\authorrunning{S. Krishnamurthy, V. Bhat, A. Jain}

\institute{
\email{krishnadots@gmail.com}
\and
Amazon Prime  Video 
\email{\{vimalb,jaabhin\}@amazon.com}}

\maketitle

\begin{abstract}
The proliferation of several streaming services in recent years has now made it possible for a diverse audience across the world to view the same media content, such as movies or TV shows. While translation and dubbing services are being added to make content accessible to the local audience,  the support for making content accessible to people with different abilities, such as the  Deaf and Hard of Hearing (DHH) community, is still lagging.   Our goal is to make media content more accessible to the DHH community by generating sign language videos with synthetic signers that are realistic and expressive.  Using the same signer for a given media content that is viewed globally may have limited appeal. Hence, our approach combines parametric modeling and generative modeling to generate  realistic-looking synthetic signers and customize their appearance based on user preferences.  We first retarget human sign language poses to 3D sign language avatars by optimizing a parametric model. The high-fidelity poses from the rendered avatars are then used to condition the poses of synthetic signers generated using a diffusion-based generative model.  The appearance of the synthetic signer is controlled by an image prompt supplied through a visual adapter. Our results show that the  sign language videos  generated using our approach have better temporal consistency and realism than signing videos generated by a diffusion model conditioned only on text prompts. We also support multimodal prompts  to allow users to further customize the signer's appearance to accommodate diversity (e.g. skin  tone, gender). Our approach is also useful for signer anonymization.
\end{abstract}

\section{Introduction}
\label{sec:intro}

The growth in media content offered by streaming services has made it possible for different types of content, such as movies, TV shows, cartoons, and even live events (\eg music concerts), to be viewed more globally by  a diverse audience.   There is also increasing support for making the content more accessible to a global audience by leveraging recent advances in large language models for dubbing and translating the content to local languages and dialects. While this makes the content more accessible to normal users, the support for making media content more accessible to people with different abilities, such as the Deaf and Hard of Hearing (DHH) community,  is still considerably lagging. These communities still rely on traditional, less expressive methods, such as captions and subtitles,  to access media content. However, captions are not a viable option for the young as well as adult audience with limited reading skills.  In contrast, sign language is a visual language that provides a more expressive alternative for  the DHH community to interpret media content and is a native language for members of the DHH community. However, the amount of media content currently far outweighs the number of human sign language experts. Moreover, given the visual nature of sign languages, some human signers are not comfortable with having their identity revealed globally in media content and  prefer anonymity.

These challenges motivate the need for streaming services to support automatic sign language generation for media content with synthetic signers, which would provide a scalable alternative for the content to be accessible by a more diverse community with different abilities. Recently, some efforts have been made to train neural architectures to generate sign language for a given linguistic input\cite{2020_stoll_signsynth, 2022_saund}. We can decouple this problem into  two main steps:  1) {\em{language2pose:}} learning a mapping from linguistic input to sign poses, and 2) {\em{pose2video:}} transferring the sign poses onto a synthetic signer (\eg an avatar or a synthetic human) to generate sign language videos.  In this paper, we primarily focus on the second step, which involves the generation of sign language videos, given the sign poses as inputs. This work is agnostic to the type of sign language and can be attached to any {\em{language2pose}} module that provides the sign poses for a language in order to create an end-to-end sign language generation system. 

We conducted a customer survey  with a group of about 100 sign language users, including professional ASL signers as well as some DHH community members. The participants were asked to indicate their preference for different interpretation methods (\eg sign language, closed captions, subtitles) that would enhance their viewing experience for different types of media and entertainment content (\eg news segment, stand-up comedy, weather forecast, shows for kids, movies etc.). We collected their feedback after viewing clips of a real human signer interpreting  different types of media content  and also asked them to compare some of these clips with video clips of an avatar signer. We learned the following  based on the feedback we gathered. 
\begin{itemize}
\item{\bf{Enhanced realism:}} While the DHH and sign language user community generally supports the use of  synthetic signers, they prefer signers that are more expressive  and human-like in appearance compared to robotic avatars.
\item{\bf{Customizability:}} While the same media content offered by the streaming services may be viewed by different users globally, having the same signer for the same media content that is viewed by a diverse DHH community across the world would limit the accessibility. Instead, users want to make the media content more accessible by having the ability to customize the appearance of the  generated signers based on the local audience or a user's personal preferences. For example, generating sign language for media content for kids by using a synthetic signer who looks like a kid instead of an adult  may make the content more appealing to the younger DHH community.  Similarly, the DHH audience in a specific geography may prefer media content with signers whose skin tone, appearance, and costume reflect the people in that local community. 
\end{itemize}
The above goals serve as the motivation for our work on generating customizable sign language videos with realistic-looking synthetic signers. Additionally, this work can also be used for sign language video anonymization that would protect the identity of the human signers who request privacy, but are willing to lend their sign poses.

\section{Design Choices and Challenges}
Different neural network models can be leveraged to generate user-configurable sign language videos with enhanced realism. Hence, we used the following requirements as the basis for our model selection. 
\begin{description}
\item {\bf{High-fidelity pose transfer:}}  Sign language videos for our media content are recorded by a human signer without the use of expensive cameras. When using these poses as conditioning inputs to generate sign language videos customized to different synthetic signers,  we want to ensure high-fidelity pose transfer that is agnostic to the customized signer, so that a given media content enhanced with sign language interpretation is interpreted in the same way by different users that view the content in different regions.  
\item {\bf{Zero-shot or few-shot customization:}} Generating sign language videos with synthetic signers that are customized to individual preferences or local communities should be achieved in a scalable manner, with little to no additional training  or fine-tuning of the models on instance-specific datasets.
A neural radiance field (NeRF) model\cite{milden20}  can be used to transfer sign language poses to a specific signer, by conditioning the model on the sign poses and training on multiple 2D views of the target signer. However, a NeRF model is typically instance-specific and needs to be retrained whenever the target signer changes, which makes it challenging to scalably customize a sign language video to different target signers.  
A conditional generative adversarial network (GAN) model has also been used to synthesize sign language videos, by training the GAN model on the sign pose sequence, conditioned on a base image of the signer\cite{2020_stoll_signsynth}. However, a conditional GAN model also needs to be retrained, in order to customize a sign language video to different target signers.

\item {\bf{Temporal consistency:}} The duration of media content may vary based on the content, such as news segments, shows for kids, stand-up comedies, and weather forecasts. We want the signing videos to be smooth  with minimal jitter, and the appearance of the generated signer to be  consistent as the signing poses change, regardless of the content duration. 

\end{description}

\subsection{Approach Overview}

 \begin{figure}
    \centering
    \includegraphics[width=12cm, height=3cm]{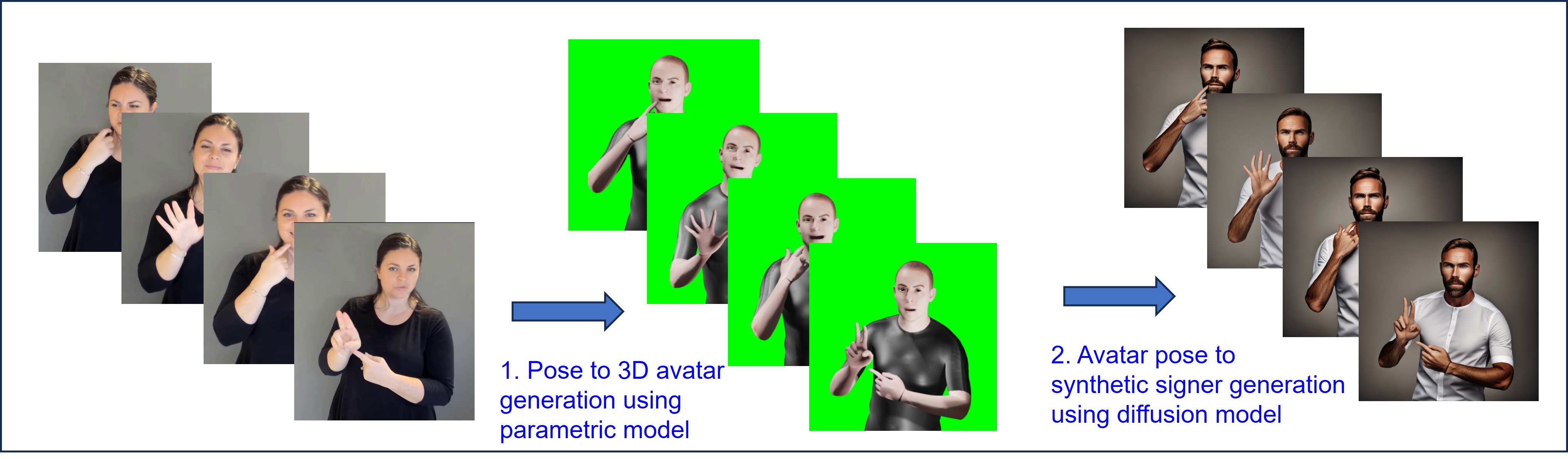}
    \caption{High-level overview of our approach}
    \label{fig:overview}
\end{figure}
Our approach for generating  sign language videos combines a parametric modeling approach to generate high-fidelity poses with a generative modeling approach to synthesize customizable, realistic-looking signers. \cref{fig:overview} provides an overview of our approach. We first extract the 2D pose sequence from a human sign language video for a media content and retarget the sign pose sequence onto a 3D avatar model by optimizing a parametric model. Retargeting to 3D avatars helps in capturing and transferring high-fidelity signing poses even when the intermediate video frames in the source video are blurry. We then leverage recent advances in diffusion-based generative models to generate user customizable sign language videos with enhanced realism without additional training, by conditioning on the poses extracted from the  rendered 3D avatars.

We experimented with video-based diffusion models (e.g. ControlVideo\cite{yzhang24_controlvideo}) to directly generate  sign language videos by  conditioning on the sign poses. However, these methods were able to generate only a few seconds of video.
Moreover,  the pose transfer was inaccurate and had very low fidelity.
Hence, the resulting sign language video was  not interpretable. Instead, our approach generates the sign language video frame-by-frame by conditioning on the pose and appearance inputs.

\begin{figure}
    \centering
    \includegraphics[width=12cm, height=3cm]{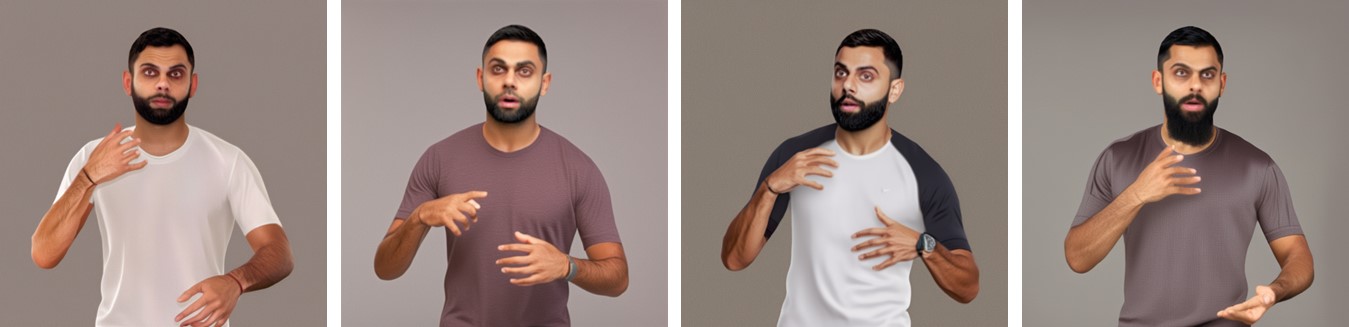}
    \caption{Frame-by-frame generation using only a text prompt to control the signer appearance  results in some inconsistency, especially for longer sign language videos. For example, the above frames were generated  using the same seed and the control prompt {\em{"a young male with beard wearing a white shirt"}}. Best viewed in color.}
    \label{fig:t2i}
\end{figure}

 One of the challenges in  performing this conditional generation on a per-frame level using text-to-image diffusion models, such as Stable Diffusion\cite{romb22}, is that it results in temporal inconsistency, since the appearance of the signer changes in each frame despite using the same seed and text prompt, as shown  in \cref{fig:t2i}, where the shirt and beard of the signer varies in the frames.  We address this challenge by conditioning on a single image of the signer, which is input to a diffusion model using a visual adapter, when generating each frame. Our results in \Cref{sec:results} show that this  generation approach using a visual adapter results in generating signing videos with realistic-looking signers that have more consistent  appearance across the frames, regardless of the duration of the media content. The signer depicted in the visual prompt can be further customized using text and other multimodal prompts to generate personalized signing videos.  

\section{Related Work} 
\label{sec:related}
In this section we discuss some of the related work in  generating sign language videos using neural models and some recent advances in diffusion-based  generative models that have influenced our approach. 

SignSynth\cite{2020_stoll_signsynth} and SignGAN\cite{2022_saund}  use a two-step neural network architecture to generate sign language videos starting from either sign glosses or text input. The former uses an auto-regressive convolutional network, while the latter uses a transformer based network to first map the glosses to a sign pose sequence.  Given the base image of a signer and the sign pose sequence as inputs, both approaches use a  conditional GAN  to generate a sequence of images showing the base signer in the input sign poses.  Additionally, SignGAN allows the generated signer to be customized to different styles. Neural Sign Reenactor\cite{2023_tze} and AnonySign\cite{2021_saund_anon} extract poses from the source sign language video and render them into photo-realistic sign language videos using a generative model that is conditioned on the target signer image. While Neural Sign Re-enactor uses a conditional GAN for rendering, AnonySign uses a conditional variational autoencoder (VAE) for generation. These methods are all trained on  conditional generative models that are target-specific and need to be retrained when the target signer changes, whereas our approach  generates sign language videos customized to different signers using zero-shot pose transfer, without additional training. 

Recent advances in diffusion models have made it possible to control the generation of images and videos using pre-trained diffusion models by conditioning on task-specific inputs. ControlNet\cite{zhang23} allows us to provide task-specific inputs, such as poses and segmentation maps, to control the generation of images using a pre-trained Stable Diffusion model\cite{romb22}. ControlVideo\cite{yzhang24_controlvideo} extends ControlNet to generate videos from text prompts by conditioning on motion sequences and uses cross-frame attention for temporal consistency. 
IP-Adapter\cite{ye23_ip} enhances a pretrained text-to-image diffusion model by using separate cross-attention modules for text and image features. This allows an image prompt to be used either individually or in  conjunction with a text prompt to control the image generation.  AnimateDiff\cite{guo23} generates short animated video clips from static images by incorporating the motion priors learned by a separate motion model from video clips into a pretrained text-to-image diffusion model.  
DiffSLVA\cite{xia23} leverages Controlnet for zero-shot, text-guided sign language video anonymization by  conditioning the pose transfer  on HED edges (Holistically-Nested Edge Detection), while the appearance of the target signer is controlled using text input to a pre-trained Stable Diffusion model. Cross-frame attention with the previous frame is used to achieve temporal consistency. While our approach also leverages ControlNet  for zero-shot pose transfer, we condition the pose transfer on canny edges and sign poses and achieve consistent signer appearance by using a single image of the  target signer as a visual prompt for generating the videos,  instead of relying on text prompt engineering to describe the appearance of the signers.

\section{Description of Approach} 
\label{sec:approach}

\begin{figure*}[t!]
    \centering
    \begin{subfigure}{0.5\textwidth}
        \centering
        \includegraphics[height=3cm]{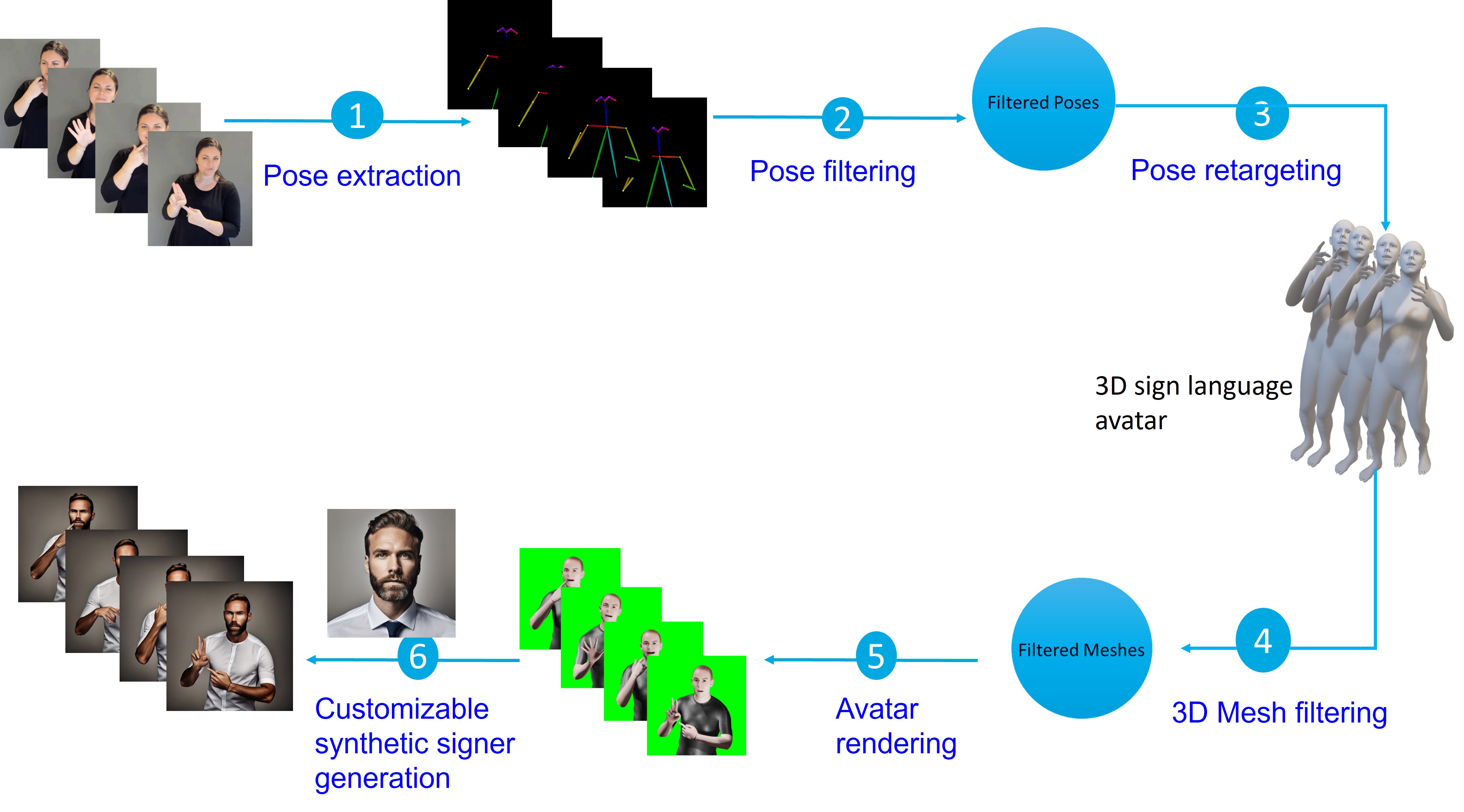}
        \caption{Pose transfer pipeline}
    \end{subfigure}%
    \hfill
    \begin{subfigure}{0.5\textwidth}
        \centering
        \includegraphics[height=3cm]{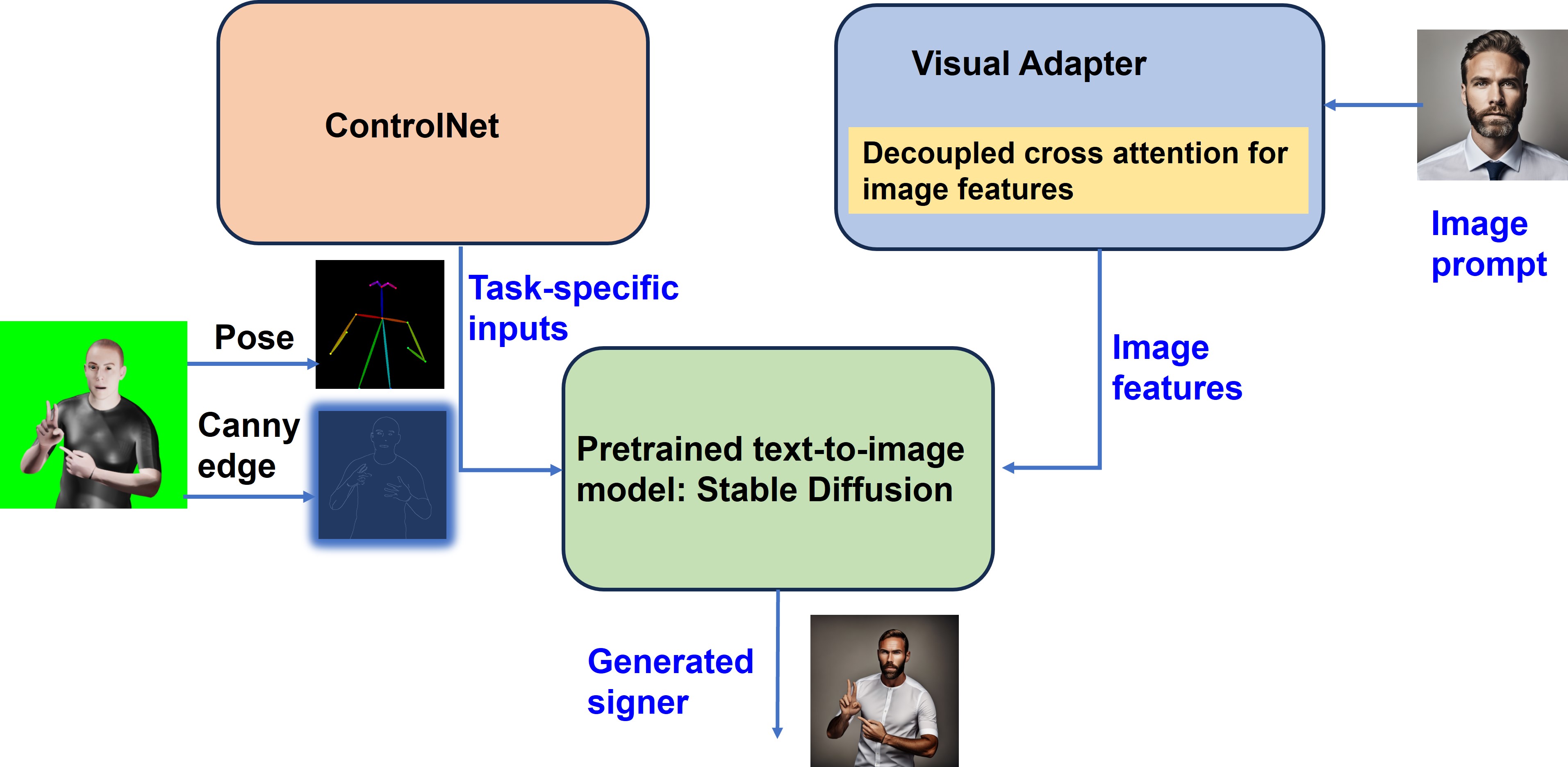}
        \caption{Architectural modules to generate synthetic human signer}
    \end{subfigure}
    \caption{Approach combining parametric and generative modeling for customizable sign language video generation with human-like synthetic signers. Best viewed in color.}
    \label{fig:approach_detail}
\end{figure*}
\cref{fig:approach_detail} shows the details of our approach for generating  sign language videos that combines a parametric modeling approach to generate high-fidelity poses with a generative modeling approach to synthesize customizable, realistic-looking signers. \cref{fig:approach_detail}a shows the sequence of steps and \cref{fig:approach_detail}b shows the architecture used by our approach to generate synthetic signing videos.  We describe each of the steps in this section.

\begin{subsection}{Pose Extraction} 
We start with the  pose extraction step, as shown in Step 1 in \cref{fig:approach_detail}a.  Given a human sign language video, we used the MediaPipe library\cite{2019_luga_mp} to extract the 2D poses of the signer that are relevant to sign language generation from each video frame. We also experimented with OpenPose\cite{2019_cao_op}, but found it to be significantly slower. The poses relevant to sign language generation include  facial, hand, finger, and upper body poses. We extracted  facial keypoints,  25 upper body keypoints, and 21 keypoints per hand, including the fingers.  MediaPipe was unable to extract some of the keypoints from blurry frames where there was rapid hand motion.  For such frames we linearly interpolated the 2D keypoints from neighboring frames.  Although we directly estimate poses from the frames of a human sign language video currently, we can also use the sign pose sequence that is output by a trained {\em{language2pose}} model, as mentioned earlier. 
\end{subsection}

\begin{subsection}{Pose Filtering} 
The frame-by-frame pose extraction causes considerable jitter in the resulting video due to keypoint shift. To smooth the poses and reduce jitter, we used the 1-\texteuro{}  (one-euro) algorithm \cite{2012_casiez}, which is a lightweight, low-pass noise filtering algorithm that is based on adaptive exponential smoothing.  This algorithm uses two hyper-parameters: the speed coefficient, $\beta$, to control the lag and minimum cutoff frequency, $f_{min}$, to control the jitter. Based on our experiments, $\beta=1.0$ and $f_{min}=0.04$ resulted in a good balance between jitter and lag for our data. 
\end{subsection}

\begin{subsection}{Pose Retargeting and Avatar Rendering}
Next, we retargeted the filtered 2D poses onto a 3D human avatar, represented by the SMPL-X parametric human body model\cite{2019_pav_smplx}. SMPL-X is defined by a function $M(\theta, \alpha, \phi)$, where  $\theta$, $\alpha$, and $\phi$ represent the  pose, shape, and facial expression parameters, respectively. We chose SMPL-X because the pose and shape parameters model the body, hands, and fingers, which are crucial for sign language generation. Pose retargeting involves fitting the SMPL-X body model to the 2D keypoints extracted from RGB images by optimizing the parameters.   We used the SMPLify-X\cite{2019_pav_smplx} algorithm for retargeting the sign poses extracted from each frame in the previous step to the SMPL-X avatar model.  

We had to address a couple of challenges when using SMPLify-X. 
First, SMPLify-X requires full body keypoints for optimization, while the  sign language videos only depict upper body, as shown in \cref{fig:overview}. Optimizing only the upper body poses resulted in invalid deformations, even though SMPLify-X uses a body pose prior for minimizing the deformation. However, the use of  MediaPipe allowed us to extract the full body keypoints with some degree of confidence  from upper body videos and we optimized only the pose and expression parameters, while freezing the shape parameters, to avoid body deformations. We also tried with OpenPose, but were not able to extract full body keypoints from upper body signing videos using OpenPose.
Second, since SMPLify-X is trained on OpenPose keypoints, whereas we used MediaPipe to extract the keypoints,  we had to first map the MediaPipe keypoints to the appropriate OpenPose keypoints. For example, MediaPipe extracts more than 400 detailed  facial keypoints, whereas OpenPose extracts only around 70 facial keypoints. So we mapped only a subset of the MediaPipe facial keypoints to the corresponding OpenPose keypoints and optimized them with SMPLify-X. 

Pose retargeting results in a 3D signing avatar corresponding to each input frame, as depicted in Step 3 in  \cref{fig:approach_detail}a. In order to further reduce jitter in the resulting video,  we filtered the vertices of the 3D avatar mesh generated for each frame using the 1-\texteuro{}   filtering approach.
Finally, each avatar mesh was rendered with appropriate texture map and lighting using Blender\cite{blender} to generate the sign language poses as a  sequence of RGB frames, as shown in Step 5 in \cref{fig:approach_detail}a. 
\end{subsection}

\begin{subsection} {Synthetic Signer Generation and Pose Transfer} 
The last step in the pipeline is to generate a synthetic human signer and transfer the sign poses from the avatar generated in the previous step  to the synthetic human signer. The appearance and pose of the synthetic signer are decoupled and controlled separately, as shown in \cref{fig:approach_detail}b. 

One way to control the appearance of the signer is by
leveraging a text-to-image diffusion model and describe the appearance of the signer using  only a text prompt. However, this resulted in inconsistent signer appearance across the generated frames despite using the same seed for each frame, while the poses were transferred with high fidelity from the avatar frames, as shown in \cref{fig:t2i}. Instead,  we leveraged a lightweight visual adapter (IP-adapter)\cite{ye23_ip} and used a single image of the target signer as an input prompt. The visual adapter extracts features  from the image prompt using a CLIP encoder\cite{clip}, which  are then used to condition the generation of the synthetic signer using a pre-trained  Stable Diffusion model\cite{romb22}. While the pre-trained  model has a cross-attention module to provide context from a text prompt, the visual adapter adds a dedicated cross-attention module to input context based on the image features  to the  U-net layers of the pre-trained  diffusion model, in order to compute the attention scores for generating the synthetic signer. This  image-based generation approach significantly improved the temporal consistency of the signer appearance across the frames. To further customize the signer appearance, a text prompt or other multimodal prompts can be optionally used in conjunction with the image prompt, as shown by the results in \Cref{sec:results}.

While the appearance is conditioned on the image prompt, in order to generate the signer with the appropriate sign pose, 
we extracted canny edges \cite{1986_canny} and poses for the face, hands, fingers, and upper body from each avatar frame rendered in the previous step and  used them as task-specific inputs to ControlNet, as shown in \cref{fig:approach_detail}b. These conditional inputs are used by Stable Diffusion for high-fidelity, frame-by-frame zero-shot pose transfer to the generated signer. The frames are then concatenated to generate the sign language video.

\end{subsection}

\section{Experimental Results}
\label{sec:results}
In this section, we present qualitative and quantitative evaluation of our approach for generating sign language videos with realistic-looking synthetic human signers. 
We also present results that show how our approach can be used to achieve our goals of generating diverse and customizable signers. 

\begin{figure}[h]
    \centering
    \includegraphics[width=12cm, height=6cm]{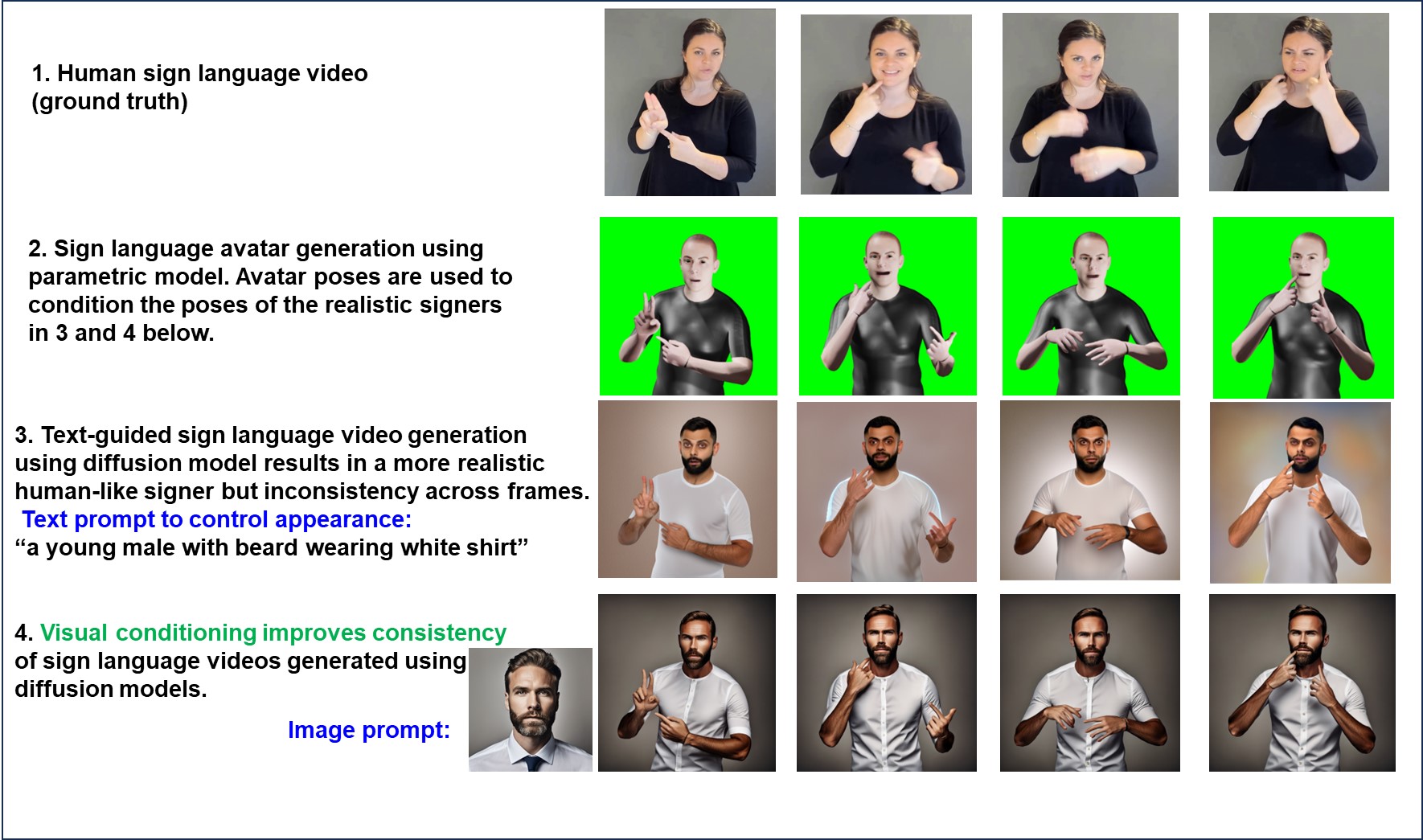}
    \caption{Improving consistency of signer appearance in the video by conditioning on an image using a visual adapter. Best viewed in color.}
    \label{fig:consistency}
\end{figure}

\subsection{Improving Temporal Consistency}
\cref{fig:consistency} compares a subset of the frames from the videos generated using different approaches with the corresponding frames in the ground-truth sign language video (shown in Row 1), which was created by a real human signer for a stand-up comedy media content. Row 2 shows the video frames from the avatar video which was rendered  after retargeting the human signer poses using the parametric approach presented in \Cref{sec:approach}.

Rows 3 and 4 in \cref{fig:consistency}  show the sign language videos with realistic-looking synthetic signers that were generated using a diffusion-based approach, while conditioning on the avatar poses supplied through ControlNet.  The video frames in Row 3 were generated using the same seed  by conditioning on the text prompt ({\em{"a young male with beard  wearing a white shirt"}}), that was input to a pre-trained text-to-image Stable Diffusion XL model, in addition to conditioning on the avatar poses from Row 2. Even though the frame-by-frame pose transfer is quite accurate, the appearance of the synthetic signer varies  across the frames, especially for long videos, resulting in a jittery video. For example, the frames in \cref{fig:t2i} and in Row 3 of \cref{fig:consistency} are from different time segments of  the same video consisting of 1380 frames and the inconsistent appearance is clearly evident.

Instead of conditioning on a text prompt, the  frames in  Row 4 were generated using our approach that uses  a visual adapter to input an image prompt  to  a pre-trained Stable Diffusion model in order to control the signer appearance.  The signer poses were conditioned on the avatar sign poses (shown in Row 2) that were input through ControlNet. Using the same visual conditioning for each frame significantly improves the temporal consistency of the signer across the video frames, resulting in lower jitter, even without explicit smoothing, instance-specific training on the target signer images, or training  on sign language content. 

\begin{figure}[t]
    \centering
    \includegraphics[width=9cm, height=2.0cm]{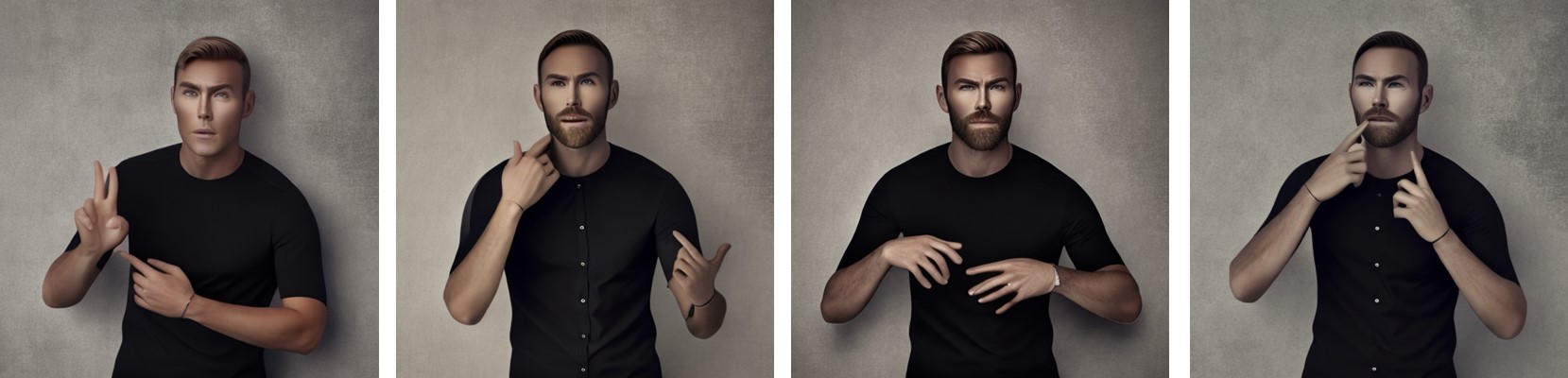}
    \caption{Personalizing the signer appearance by fine-tuning the diffusion model on a few images of the target signer using Dreambooth.}
    \label{fig:dreambooth}
\end{figure}

\subsection{Personalized Signer Generation using Fine-Tuning}
\label{sec:db}
In the previous section, we compared two diffusion-based approaches for generating synthetic sign language videos, one of which used a text prompt while the other used a visual conditioning approach to  control the appearance of the synthetic signer generated by  a pre-trained Stable Diffusion model. In both cases, the diffusion-based generative model was used in inference mode for zero-shot pose transfer. It  was not trained or fine-tuned on  the target signer images. 

In this section, we present results of a third approach for personalizing the appearance of the synthetic signer, which involves  fine-tuning the pretrained Stable Diffusion model with just a few images of the target signer. Our goal was to determine if fine-tuning improves the generation quality. To do this, we  used
DreamBooth\cite{ruiz23}, which is a method for fine-tuning pre-trained, large-scale, text-to-image diffusion models with just a few images of a personalized subject, by associating a unique identifier with that subject. The fine-tuned model can then be used to generate the subject in personalized contexts by embedding the unique identifier in the text prompt provided as a conditional input for the generation. We fine-tuned StableDiffusion v1.5 using DreamBooth with 5 frames from the synthetic  signer video shown in Row 4 of \cref{fig:consistency} and associated these images with a random unique identifier, {\em{"<sks> Edward"}} to personalize the signer.  Then we used this fine-tuned model to generate the sign language video by conditioning the signer appearance on a simple personalized text prompt which includes the  identifier associated during fine-tuning:  {\em{"<sks> Edward wearing a black shirt"}}. 
Similar to the results in \cref{fig:consistency}, zero-shot pose transfer to this personalized signer was achieved by supplying the canny edges and poses extracted from the avatar video  as task-specific inputs through a pre-trained ControlNet model. 

\cref{fig:dreambooth} shows the results of the fine-tuned model. We see that the model that was fine-tuned on just 5 images of the target signer and conditioned on a simple personalized text prompt is able to generate a signer whose appearance closely matches the target signer (shown in Row 4 of \cref{fig:consistency}).  However,  the appearance of the signer in \cref{fig:dreambooth} is not consistent across frames, despite personalization. For example,  there is no beard in the first frame although all the 5 images used for fine-tuning had a beard. In comparison, our approach of supplying a  single image of the target signer through a visual adapter to control the signer appearance  results in a sign language video with better temporal consistency even without fine-tuning,  as shown in Row 4 of  \cref{fig:consistency}.

\begin{table}
    \centering
     \caption{Quantitative evaluation of sign language videos generated using parametric and generative models.}
    \begin{tabular}{||c|c|c|c|c||}
       \hline \hline
       &{\bf{Sign Language Video Generation}} & {\bf{Structural}} & {\bf{Directional}} & {\bf{Avg. FID $\downarrow$}} \\
       &{\bf{Approach}} & {\bf{Similarity$\uparrow$}} & {\bf{Similarity$\uparrow$}} & \\
       \hline \hline
       1 & Avatar Video  &  {\bf{0.812}} & 0.185 & 176.035 \\
       &(parametric + pose smoothing)  &&& \\
       \hline
       2 & Pre-trained diffusion model  & 0.553 & 0.183 & 158.525 \\
       &(only text prompt) &&& \\
       \hline
       3 & Pre-trained diffusion model  &  0.769 & {\bf{0.194}} & 146.490 \\
       &(image prompt+visual adapter)  &&& \\
       \hline 
       4 & Fine-tuned diffusion model   &  0.668 & 0.169 & {\bf{130.896}} \\
       &(only text prompt) &&& \\
       \hline 
    \end{tabular}
   
    \label{tab:ssim}  
\end{table}

\subsection{Quantitative Evaluation}
\cref{fig:consistency} presented a qualitative comparison of the sign language videos with synthetic signers generated using  parametric and diffusion-based generative approaches. 
In this section, we use different metrics to quantitatively evaluate the temporal consistency, alignment with prompt semantics, and the realism of the synthetic sign language videos generated using the different models. 

1) {\bf{Structural Similarity (SSIM):}} We computed the structural similarity (SSIM) between consecutive pairs of frames to measure the jitter in the generated video. In \Cref{tab:ssim}   we have reported the average SSIM computed over the entire video having 1380 frames. Higher the SSIM value, lower is the jitter and hence, better is the consistency or smoothness of the generated video.  
We see that the video output by our generative approach based on visual conditioning (frames shown in Row 4 of \cref{fig:consistency}) has an SSIM value of 0.769. 
Fine-tuning the  diffusion model with a few images of the target signer and conditioning the generation on a personalized text prompt, as described in \Cref{sec:db}, results in SSIM=0.668, while
conditioning a  pre-trained diffusion model on a text prompt without any fine-tuning results in  SSIM=0.553. These results show that our approach of visually conditioning the signer generation on just a single signer image using a pre-trained model with a visual adapter, results in lower jitter compared to the text conditioned models and outperforms even the fine-tuned model. Furthermore, by comparing the two text-conditioned approaches we infer that fine-tuning on a few target signer images helps lower the jitter. 

Finally, we see that even without any pose smoothing or frame interpolation, the video generated using our visual conditioning approach (with SSIM=0.769) has only slightly higher jitter than the avatar video (with SSIM=0.812) that is rendered after explicit pose smoothing using the 1-\texteuro{} algorithm.  To summarize, conditioning the signer generation  on just a single image of the target signer supplied through a visual adapter improves the temporal consistency of the  sign language video,  without the need for fine-tuning on additional signer images, complex prompt engineering, or explicit pose smoothing. The consistency of the generated video may be further improved using frame interpolation methods.

2) {\bf{Directional Similarity (DS):}} The transfer of sign poses from a human signer video to a synthetic signer or avatar video effectively results in a change in the signer appearance while preserving the poses. Hence, this can be regarded as a type of  style transfer or editing operation where some aspects of the original content are modified and one way to quantitatively evaluate the generated sign video is by measuring how well it conforms to the editing instructions. Given a pair of images $(i_s, i_t)$ and text descriptions $(d_s, d_t)$, where  $d_s$ describes the source image $i_s$ and $d_t$ describes the target image $i_t$ resulting from editing $i_s$, the directional similarity measures how well the change in the edited image aligns with the editing semantics. To compute this, we encoded the images using the CLIP image encoder\cite{clip}, $E_i$, and the text descriptions using the CLIP text encoder, $E_d$. We then computed DS using cosine similarity between the encoded image and text, as follows: $DS = cosineSim((E_i(i_t) - E_i(i_s)),(E_d(d_t) - E_d(d_s)))$. 

We computed the frame-wise DS between the human sign language video, that is regarded as the source (Row 1 in \cref{fig:consistency}) and the synthetically generated signer video, regarded as the target. The average DS over the entire video for each of the generative methods is reported in \Cref{tab:ssim}. To compute the DS, a simple high-level description was used for all the frames of the real human source video: $d_s$ = {\em{"a female sign language signer"}}. Similarly we used a high-level description for all the frames of the generated video to match the appearance of the synthetic signer. For \eg the video frames generated by the visual prompt based diffusion method (shown in Row 4 in \cref{fig:consistency}) were described by $d_t$ = {\em{"a young male sign language signer with a beard wearing a white shirt"}}. From \Cref{tab:ssim} we see that  the visual adapter based method has the highest directional similarity. Even though this method was conditioned only on a single image prompt and did not use any text prompts like the other two diffusion-based methods in the table, it is able to generate a signer whose appearance is more consistently aligned with the semantics of the style editing descriptions that indicate that the  sign poses are transferred from a  female signer (source) to a bearded, male signer (target).

3) {\bf{Frechet Inception Distance (FID):}} To compare the realism and visual quality of the  generated videos,   we computed the FID score\cite{2018_lucic}  with respect to the human sign language video (Row 1 of \cref{fig:consistency}) using the FID implementation in the {\tt{torchmetrics}} library. A lower FID score signifies higher visual quality and  realism of the signer appearance.  Based on the FID scores in \Cref{tab:ssim}, the avatar video has the highest FID score=176.035, as expected, since the avatar has the least human-like appearance (see frames in Row 2 of \cref{fig:consistency}). In comparison, the video generated by our visual conditioning approach has a lower FID score=146.49, indicating better visual quality and realism compared to the avatar video and the video generated using the pre-trained diffusion model conditioned on a text prompt (FID=158.525).  Fine-tuning the diffusion model with a few images of the target signer helps in further improving the visual quality of the  signer  (see \cref{fig:dreambooth}) and results in the lowest FID value=130.896.
Evaluation of the synthetic signing videos using human subjects will be part of future work.

\subsection{Ablation}

\begin{figure*}[t]
    \centering
    \begin{subfigure}{0.4\textwidth}
        \centering
        \includegraphics[height=2cm]{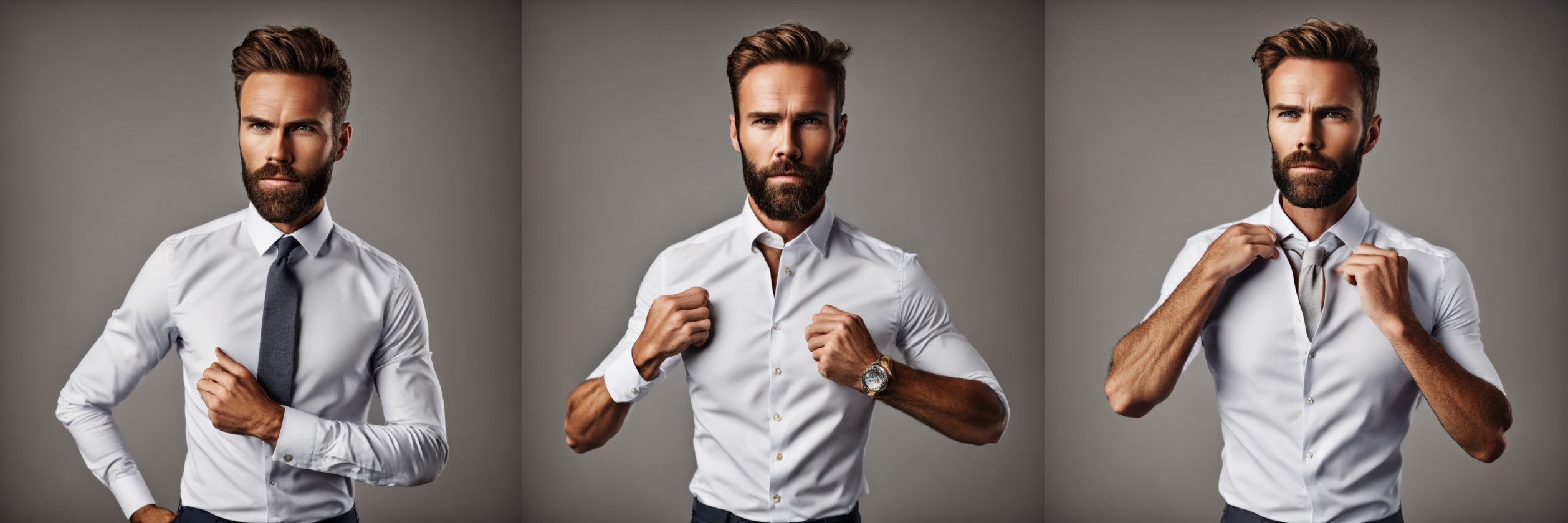}
        \caption{Conditioning only on poses}
    \end{subfigure}%
    \hfill
    \begin{subfigure}{0.5\textwidth}
        \centering
        \includegraphics[height=2cm]{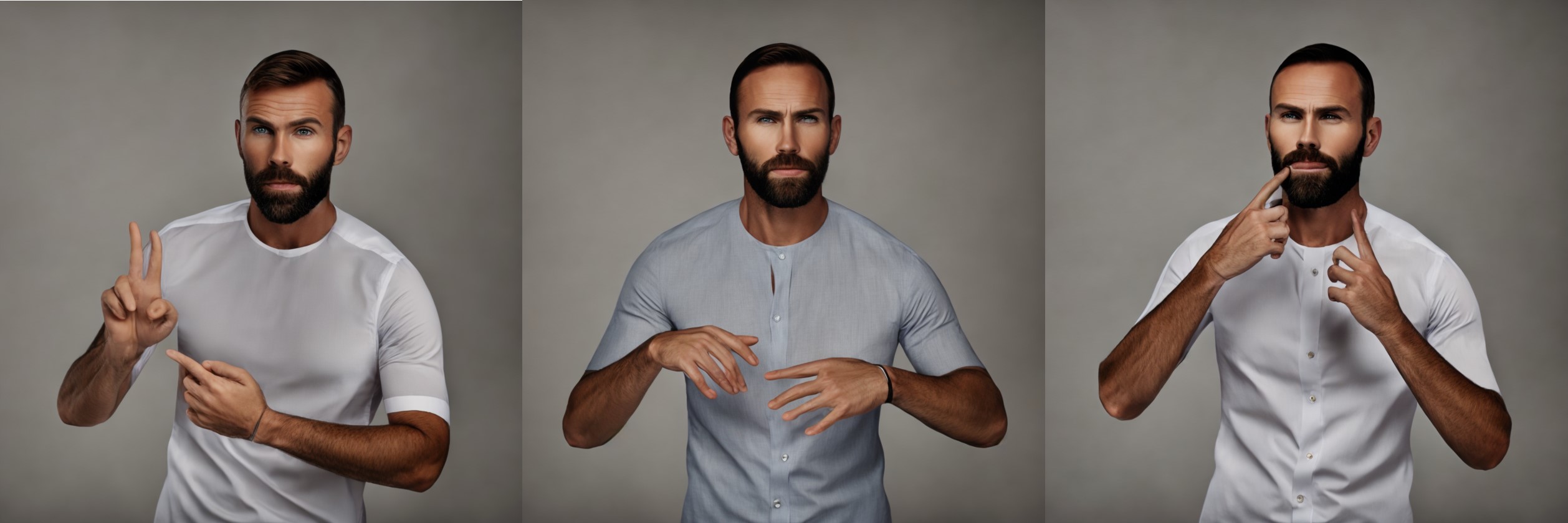}
        \caption{Conditioning on canny edge and depth map}
    \end{subfigure}
    \caption{Ablation results for pose transfer}
    \label{fig:ablation_pose_cond}
\end{figure*}

\cref{fig:ablation_pose_cond}, shows the impact of using alternative  task-specific inputs to ControlNet for conditioning the signer generation.
From \cref{fig:ablation_pose_cond}a, we see that if we use only the poses extracted from the avatar video  as the conditional inputs, the resulting poses are incorrect and in some cases, do not even represent sign language poses. On the other hand,   \cref{fig:ablation_pose_cond}b shows that conditioning on  a combination of canny edge and depth map transfers the poses  correctly, but results in inconsistent signer appearance across different frames.  We found that conditioning on a combination of canny edges and pose input results in more accurate pose transfer to the generated signer,  as shown in \cref{fig:consistency}, with the contours of the body parts provided by the canny edge detection contributing to the improvement.

\subsection{Customization of Signer Appearance}
In order to achieve our goal of allowing users to customize the appearance of the signer based on their preferences, our approach supports multimodal prompts.
\begin{figure}
    \centering
    \includegraphics[width=11cm, height=5cm]{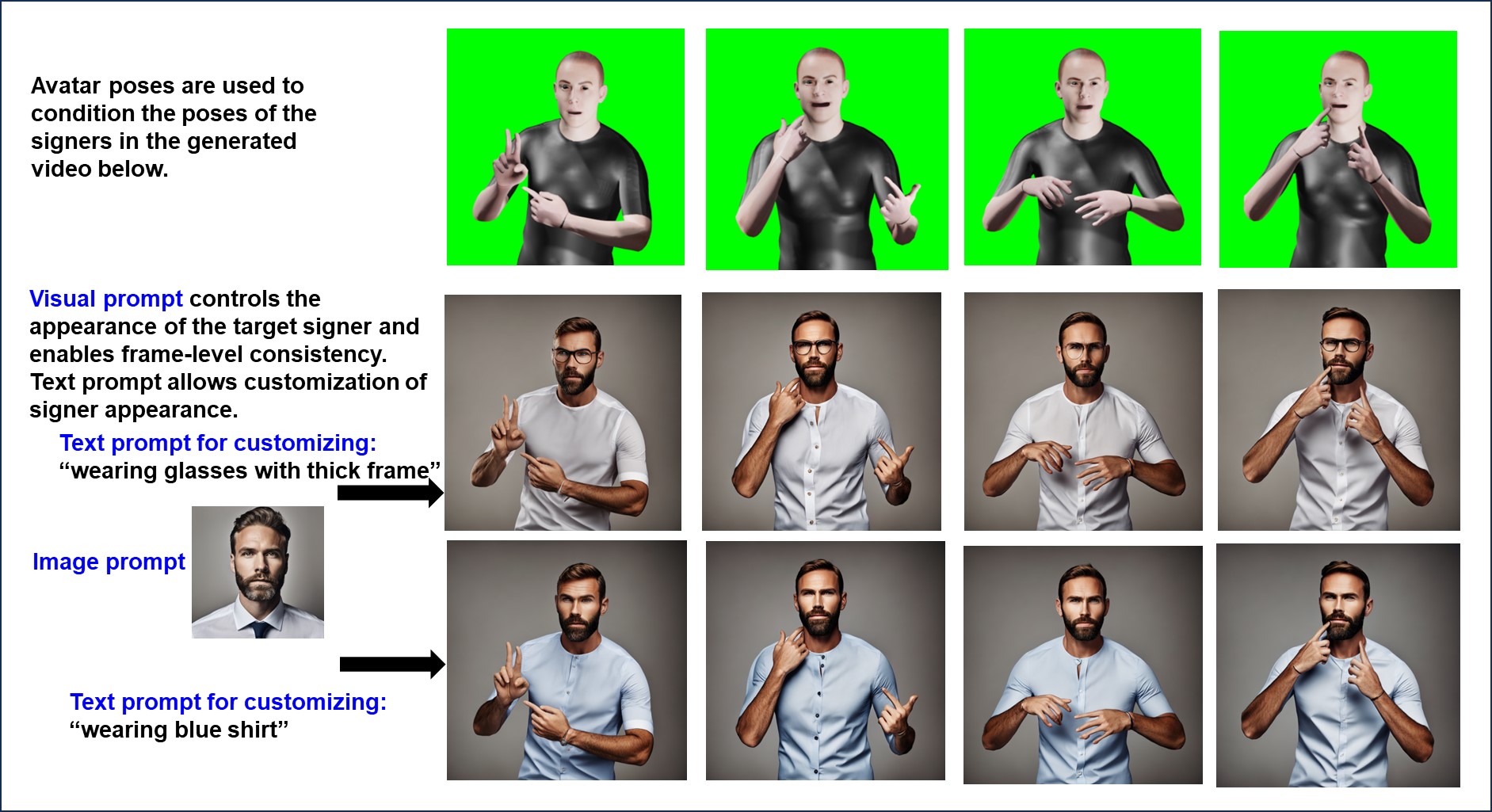}
    \caption{Customizing the appearance with multimodal prompts. Best viewed in color.}
    \label{fig:custom}  
\end{figure}
As stated earlier, we input the canny edge and pose inputs through ControlNet to condition the signer poses  and an image prompt through the visual adapter to control the base appearance of the synthetic signer. Additionally, a text prompt can be supplied to Stable Diffusion to further customize the appearance of the target signer.  \cref{fig:custom}  shows how a  text prompt in conjunction with an image prompt is used to generate sign language videos with the same signer wearing glasses in Row 2  and wearing a blue-colored shirt in Row 3, instead of a white-colored shirt.

\subsection{Signer Diversity}
\begin{figure}[t]
    \centering
    \includegraphics[width=11cm, height=5cm]{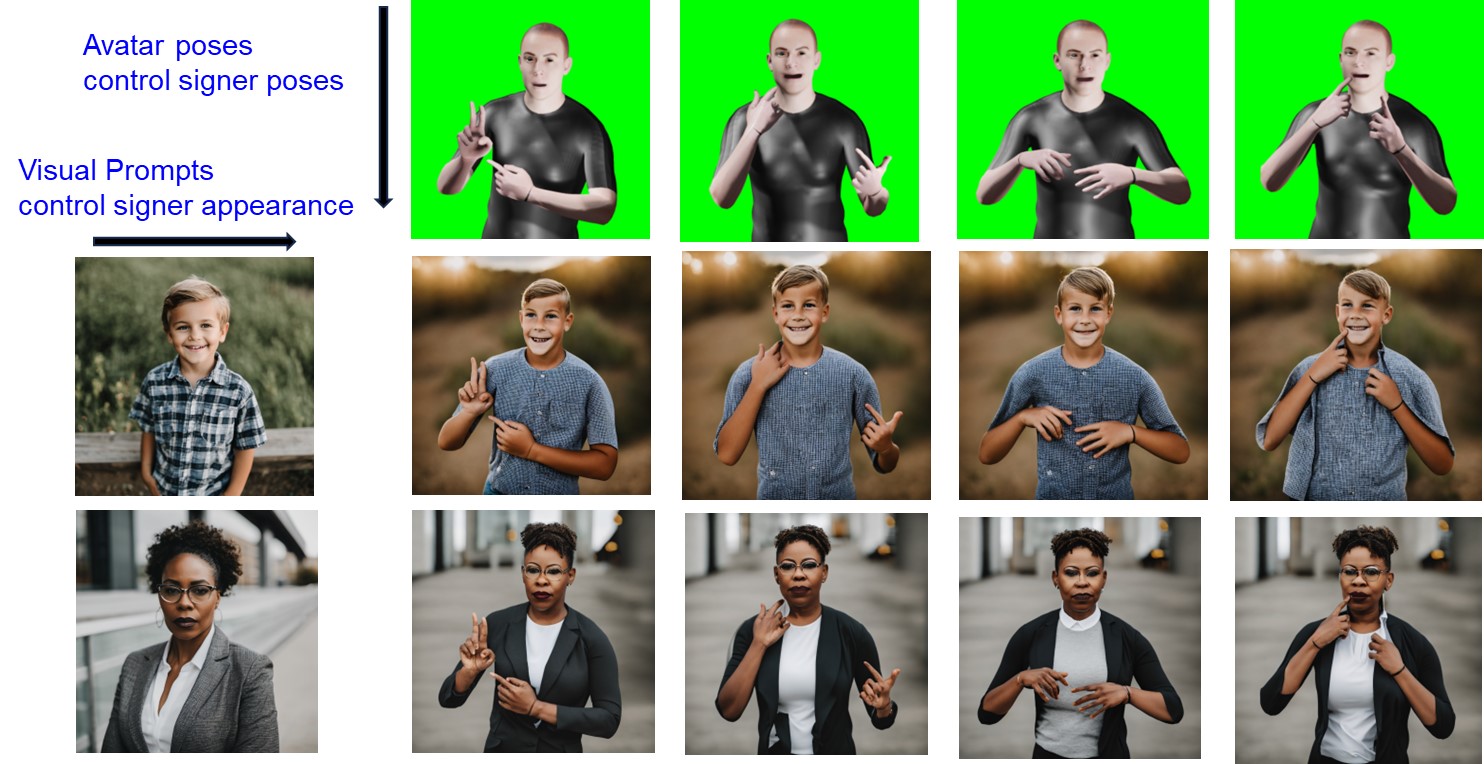}
    \caption{Zero-shot signer diversity using visual prompts}
    \label{fig:signer_diversity}  
\end{figure}

\cref{fig:signer_diversity} shows how we generated sign language videos for the same media content using diverse synthetic signers, in order  to make the content more accessible to different communities.  Each row shows a subset of frames from the sign language video for the same media content using a different synthetic, but realistic-looking signer. 
First, we used simple text prompts to generate a base image of synthetic human signers that are diverse in age group, demography, gender, and race using a Stable Diffusion model. \cref{fig:signer_diversity} shows two examples of these synthetically generated target signers in the leftmost column, with a young boy in Row 2 and an African-American woman in Row 3. Then this base image is input as a visual prompt through the visual adapter, in order to control the appearance of the signer in the generated signing video. The avatar poses, as shown in the topmost row,  are used as conditioning inputs for frame-by-frame zero-shot transfer of the poses to the synthetic signers. These diverse synthetic  signers can be used to sign the video content targeting different DHH communities. For example, the young boy as a synthetic signer can be used for signing kids shows and cartoons.

\section{Conclusions}
\label{sec:concl}
In this paper, we presented an approach that combines the strengths of parametric models and generative models to generate synthetic sign language videos with the goal of making  media content more accessible to a diverse and inclusive community. 
We generated expressive and realistic-looking synthetic signers and customized their appearance using diffusion-based generative models, by conditioning on just a single image of the target signer, which was  supplied through a lightweight visual adapter. We transferred the sign poses from a human sign language video to  a 3D human avatar by optimizing a parametric model and used the high-fidelity poses from the rendered avatars to condition the poses of the synthetic signers in a zero-shot manner. Our results show that our visual conditioning approach 
for generating sign language videos, with decoupled control of the pose and appearance, 
results in a more realistic and consistent signer appearance across the video frames, compared to a text-to-image generation approach.
Our approach can also be used for sign language video anonymization.
{\small{The code is available at {\em{https://github.com/athenas-lab/DiffSign}}}}.
\bibliography{main}
\bibliographystyle{plain}

\end{document}